\pdfoutput=1

\documentclass[11pt]{article}

\usepackage[final]{acl}

\usepackage{times}
\usepackage{latexsym}

\usepackage[T1]{fontenc}

\usepackage[utf8]{inputenc}

\usepackage{microtype}

\usepackage{inconsolata}

\usepackage{graphicx}

\title{Factuality or Fiction? \\ Benchmarking Modern LLMs on Ambiguous QA with Citations}

\author{Maya Patel \\
  Independent Scholar \\
  \texttt{pmaya5279@gmail.com} \\\And
  Aditi Anand \\
  Independent Scholar \\
  \texttt{anand113@gmail.com}}

\begin{document}
\maketitle
\begin{abstract}
Benchmarking modern large language models (LLMs) on complex and realistic tasks is critical to advancing their development. In this work, we evaluate the factual accuracy and citation performance of state-of-the-art LLMs on the task of Question Answering (QA) in ambiguous settings with source citations. Using three recently published datasets—DisentQA-DupliCite, DisentQA-ParaCite, and AmbigQA-Cite—featuring a range of real-world ambiguities, we analyze the performance of two leading LLMs, GPT-4o-mini and Claude-3.5. Our results show that larger, recent models consistently predict at least one correct answer in ambiguous contexts but fail to handle cases with multiple valid answers. Additionally, all models perform equally poorly in citation generation, with citation accuracy consistently at 0. However, introducing conflict-aware prompting leads to large improvements, enabling models to better address multiple valid answers and improve citation accuracy, while maintaining their ability to predict correct answers. These findings highlight the challenges and opportunities in developing LLMs that can handle ambiguity and provide reliable source citations. Our benchmarking study provides critical insights and sets a foundation for future improvements in trustworthy and interpretable QA systems.
\end{abstract}

\section{Introduction}
Large language models (LLMs) have revolutionized numerous tasks, including question answering (QA) \cite{openai2023, openai2023gpt4, anthropic, touvron2023llama, mpt_2023, almazrouei2023falcon}, summarization \cite{liu2024learningsummarizelargelanguage}, and creative content generation \cite{chakrabarty-etal-2023-creative}. Their ability to produce human-like text has driven adoption across diverse fields such as education \cite{han-etal-2024-llm, chiang-etal-2024-large, carpenter-etal-2024-assessing, shaier2024malamutemultilingualhighlygranulartemplatefree, helcl-etal-2024-teaching-llms, macias-etal-2024-empowering}, healthcare \cite{qin-etal-2024-enhancing, pandey2024advancinghealthcareautomationmultiagent, shaier-etal-2024-comparing, shaier-etal-2023-emerging, zhang-etal-2024-llm-based, garcia-ferrero-etal-2024-medmt5}, and scientific research \cite{hsu-etal-2024-chime}. However, these models are often criticized for generating hallucinations—outputs that, while plausible, are factually inaccurate or unsupported by evidence \cite{rawte-etal-2023-troubling, shaier-etal-2023-stochastic, semnani-etal-2023-wikichat, li-etal-2023-halueval, dziri-etal-2022-origin}. This issue poses a significant challenge, particularly for applications requiring reliability and accountability. Addressing hallucinations is crucial to enhancing the trustworthiness and practical utility of LLMs.

To improve the factuality of LLM outputs, researchers have explored various strategies. These include fine-tuning models on high-quality datasets, leveraging retrieval-augmented generation \cite{xia-etal-2024-rule}, and employing advanced prompting techniques \cite{zhang-gao-2023-towards}. A notable direction involves developing benchmark datasets designed to assess both the factual correctness of LLM-generated content and their ability to cite credible sources. These dual-purpose benchmarks allow researchers to evaluate not just the accuracy of answers but also the reliability of the evidence provided. Recent datasets like DisentQA-DupliCite, DisentQA-ParaCite, and AmbigQA-Cite \cite{shaier-etal-2024-adaptive} focus on QA tasks featuring ambiguous questions—where multiple valid answers may exist—and require source citation, making them critical tools for advancing LLM evaluation frameworks.

To the best of our knowledge, this is the first work to evaluate the performance of state-of-the-art LLMs, GPT-4o-mini and Claude-3.5, on ambiguous QA tasks using these datasets. Prior research has primarily focused on simpler QA scenarios or synthetic datasets, often emphasizing factuality without systematically addressing ambiguity or citation accuracy. By contrast, this study analyzes how modern LLMs handle real-world ambiguities while meeting the dual demands of accurate answer generation and reliable source citation. Our findings contribute to the growing body of research on benchmarking LLMs under more realistic and challenging conditions.

Our results reveal that both GPT-4o-mini and Claude-3.5 exhibit strengths and weaknesses. Larger, more advanced models consistently predict at least one correct answer in ambiguous contexts, demonstrating their capacity to process complex scenarios. However, they struggle to address cases involving multiple valid answers, frequently defaulting to oversimplified responses. Citation performance remains particularly weak, with citation accuracy effectively at 0 across all models evaluated. Importantly, we show that conflict-aware prompting, previously introduced in other studies \cite{shaier-etal-2024-adaptive}, can significantly improve model performance in handling multiple valid answers and generating accurate citations for these datasets.

By evaluating these state-of-the-art LLMs, we provide insights into the challenges and opportunities associated with building models capable of navigating ambiguity and producing reliable citations. This benchmarking study underscores the importance of datasets that simulate real-world complexities and highlights the need for continuous innovation in techniques to improve factuality and interpretability in LLMs.

\section{Related Work}  

\subsection{Hallucinations in LLMs}  
LLMs are known for their ability to generate human-like text across a variety of domains \cite{openai2023, openai2023gpt4, anthropic}. However, they are prone to hallucinations—outputs that are factually incorrect or unsupported by evidence \cite{rawte-etal-2023-troubling, shaier-etal-2023-stochastic, shaier-etal-2024-desiderata, semnani-etal-2023-wikichat, li-etal-2023-halueval, dziri-etal-2022-origin}. This issue has been extensively documented in tasks such as summarization \cite{cao-etal-2022-hallucinated, li-etal-2024-better, zhao-etal-2020-reducing, ramprasad-etal-2024-analyzing}, translation \cite{wang-sennrich-2020-exposure, benkirane-etal-2024-machine}, and QA \cite{sadat2023delucionqadetectinghallucinationsdomainspecific, sun-etal-2024-benchmarking, shaier-etal-2024-say, ramakrishna-etal-2023-invite, jiang-etal-2024-large}. Hallucinations can arise due to limitations in training data, the inherent uncertainty of certain queries, or a model's usage of their parametric knowledge which can be outdated. Recent studies have focused on identifying the root causes of hallucinations and proposing solutions, such as dataset modification, architectural changes, and post-hoc verification systems. Despite these advances, the problem persists, particularly in complex tasks like ambiguous QA, where the boundaries between factuality and interpretation are less clear.  

\subsection{Factuality Benchmarks for LLMs}  
Benchmarking LLMs on factual accuracy has become a critical research area to address their reliability. Traditional benchmarks such as SQuAD \cite{rajpurkar2018knowdontknowunanswerable}, Natural Questions \cite{kwiatkowski-etal-2019-natural}, and TriviaQA \cite{joshi-etal-2017-triviaqa} focus on evaluating correctness in answering well-defined queries. However, these benchmarks do not account for ambiguity or the necessity of source citation, leaving a gap in assessing models’ real-world applicability. Recent benchmarks like FEVER \cite{thorne-etal-2018-fever} and TabFact \cite{chen2020tabfactlargescaledatasettablebased} introduce the notion of evidence-based evaluation but remain limited to binary verification tasks. Newer datasets, such as DisentQA-DupliCite, DisentQA-ParaCite, and AmbigQA-Cite, have shifted focus towards testing factuality in ambiguous contexts with an emphasis on citation accuracy. These datasets enable researchers to assess whether models can generate accurate answers while appropriately attributing their sources, offering a more holistic view of model performance.  

\subsection{Ambiguous QA Datasets and Citation Evaluation}  
Ambiguous QA has gained attention as an area where traditional models often fall short \cite{min-etal-2020-ambigqa}. Questions in this domain admit multiple valid interpretations or answers, making them inherently more challenging for LLMs. AmbigQA \cite{min-etal-2020-ambigqa} is one of the pioneering datasets in this space, focusing on generating disambiguated answers along with contextually appropriate interpretations. Building on this, datasets like DisentQA-DupliCite and DisentQA-ParaCite introduce the requirement for source citations, further complicating the task and providing a more realistic test of LLM capabilities. Citation evaluation has become a critical aspect of these benchmarks, with metrics that assess the relevance, correctness, and trustworthiness of cited sources still being developed. These datasets have proven instrumental in understanding how LLMs handle complex, real-world ambiguities while maintaining factual rigor.  

\subsection{Advancements in Prompting Strategies}  Prompt engineering has emerged as a powerful technique for enhancing LLM performance on specific tasks \cite{sahoo2024systematicsurveypromptengineering, chen2024unleashingpotentialpromptengineering, white2023promptpatterncatalogenhance}. Early works focused on simple task-specific prompts, while recent approaches like chain-of-thought prompting \cite{wei2023chainofthoughtpromptingelicitsreasoning} and self-consistency prompting \cite{wang2023selfconsistencyimproveschainthought} have shown substantial improvements in reasoning and accuracy. Conflict-aware prompting, introduced in previous research \cite{shaier-etal-2024-adaptive}, is particularly relevant for ambiguous QA tasks. It explicitly guides models to consider conflicting or multiple valid answers, improving both disambiguation and response quality. While much of the prompting literature has centered on improving factuality and reasoning, its application to citation accuracy remains underexplored. This work builds on these advancements by evaluating conflict-aware prompting on the challenging datasets introduced above, shedding light on its effectiveness for modern LLMs like GPT-4o-mini and Claude-3.5.

\section{Methodology}  
\subsection{Datasets}  
We evaluate model performance using three recently published datasets: \textbf{DisentQA-DupliCite}, \textbf{DisentQA-ParaCite}, and \textbf{AmbigQA-Cite}. These datasets are designed to test QA systems in ambiguous settings where multiple valid answers may exist and require accurate source citations. These datasets provide a challenging evaluation framework for LLMs, capturing real-world ambiguities and emphasizing the importance of citation reliability.  

\subsection{LLMs Evaluated}  
We benchmark two state-of-the-art LLMs:  
- \textbf{Claude-3.5-Haiku-20241022}, one of the latest iteration of Anthropic's Claude series.  
- \textbf{GPT-4o-Mini-2024-07-18}, a compact but high-performing version of OpenAI's GPT-4.  

To contextualize their performance, we compare these models to five LLMs previously evaluated on similar tasks:  
- \textbf{Llama-2-7B Chat} \cite{touvron2023llama}
- \textbf{Llama-2-13B Chat} \cite{touvron2023llama}
- \textbf{Llama-2-70B Chat Instruct} \cite{touvron2023llama}
- \textbf{MPT-7B} \cite{mpt_2023}
- \textbf{Falcon-7B Instruct} \cite{almazrouei2023falcon}

This comparative analysis enables us to understand the progress made by Claude-3.5 and GPT-4o-mini in addressing ambiguous QA with citations.  

\subsection{Evaluation Metrics}  
We adopt the evaluation metrics introduced in the original datasets' paper, to comprehensively assess performance on ambiguous QA tasks with source citations:  

- \textbf{Accuracy at K (Acc\_K)}: This metric evaluates a model’s ability to generate at least $K$ correct answers from the gold set of answers. For a given question with gold answers \{“X”, “Y”, “Z”\} and model outputs \{“X”, “Y”\}, the scores would be Acc\_1 = 1, Acc\_2 = 1, and Acc\_3 = 0. This captures a model’s ability to generate diverse, valid responses.  

- \textbf{Citation Accuracy (A\_C)}: This metric measures the accuracy of citation strings associated with correct answers. For instance, if the gold answers are \{“According to Document X, the answer is X1”, “According to Document Y, the answer is Y1”\} and the generated outputs are \{“According to Document X, the answer is X1”, “According to Document Z, the answer is Y1”\}, the score would be 0.5.  

These metrics provide a nuanced understanding of a model's performance in handling ambiguous contexts and reliably citing sources, as described in prior work.  

\subsection{Experimental Setup}  
Our experiments evaluate the performance of the selected LLMs under two prompting strategies:  
1. \textbf{Standard Prompting}: The baseline approach, where models are queried without additional guidance on resolving ambiguity or citing sources.  
2. \textbf{Conflict-Aware Prompting}: A method proposed in the original datasets' paper that explicitly encourages models to identify and reconcile conflicting information, generating distinct answers and accurate citations.  

We compare model performance across these setups using the metrics described above, focusing on their ability to address ambiguous QA scenarios and produce reliable citations. All experiments are conducted using the default API configurations for Claude-3.5 and GPT-4o-mini, ensuring consistent evaluation conditions.  

\begin{table*}[h!]
\centering 
\caption{Results for AmbigQA-Cite}
\begin{tabular}{|l|c|c|c|}
\hline
\textbf{Model} & \textbf{A\_1} & \textbf{A\_2} & \textbf{A\_C} \\ \hline
\multicolumn{4}{|c|}{\textbf{Standard Prompt}} \\ \hline
Llama-7B & 54.8 & 2.1 & 0.0 \\ \hline
Llama-13B & 48.3 & 3.2 & 0.0 \\ \hline
Llama-70B & 54.8 & 4.3 & 0.0 \\ \hline
MPT-7B & 50.5 & 0.0 & 0.0 \\ \hline
Falcon-7B & 30.1 & 1.0 & 0.0 \\ \hline
GPT-4o-mini-2024-07-18 & 100 & 37.63 & 0.0 \\ \hline
Claude-3.5-Haiku-20241022 & 100 & 54.84 & 0.0 \\ \hline
\multicolumn{4}{|c|}{\textbf{Conflict-Aware Prompt}} \\ \hline
Llama-7B & 62.3 & 21.5 & 34.4 \\ \hline
Llama-13B & 67.7 & 22.5 & 36.5 \\ \hline
Llama-70B & 74.1 & 35.4 & 48.3 \\ \hline
MPT-7B & 46.2 & 9.6 & 21.5 \\ \hline
Falcon-7B & 39.7 & 5.3 & 16.6 \\ \hline
GPT-4o-mini-2024-07-18 & 100 & 52.69 & 0.0 \\ \hline
Claude-3.5-Haiku-20241022 & 100 & 61.19 & 0.0 \\ \hline
\end{tabular}
\label{AmbigQA-Cite}
\end{table*}

\begin{table*}[h!]
\centering 
\caption{Results for DisentQA-ParaCite}
\begin{tabular}{|l|c|c|c|}
\hline
\textbf{Model} & \textbf{A\_1} & \textbf{A\_2} & \textbf{A\_C} \\ \hline
\multicolumn{4}{|c|}{\textbf{Standard Prompt}} \\ \hline
Llama-7B & 69.6 & 7.3 & 0.0 \\ \hline
Llama-13B & 71.6 & 4.3 & 0.0 \\ \hline
MPT-7B & 65.3 & 0.3 & 0.0 \\ \hline
Falcon-7B & 50.6 & 7.6 & 0.0 \\ \hline
GPT-4o-mini-2024-07-18 & 100 & 0.0 & 0.0 \\ \hline
Claude-3.5-Haiku-20241022 & 100 & 60.0 & 0.0 \\ \hline
\multicolumn{4}{|c|}{\textbf{Conflict-Aware Prompt}} \\ \hline
Llama-7B & 74.3 & 56.0 & 59.0 \\ \hline
Llama-13B & 77.0 & 58.0 & 60.6 \\ \hline
MPT-7B & 54.3 & 26.6 & 32.0 \\ \hline
Falcon-7B & 54.3 & 19.3 & 30.3 \\ \hline
GPT-4o-mini-2024-07-18 & 100 & 88.0 & 0.0 \\ \hline
Claude-3.5-Haiku-20241022 & 100 & 86.3 & 0.0 \\ \hline
\end{tabular}
\label{DisentQA-ParaCite}
\end{table*}

\begin{table*}[h!]
\centering 
\caption{Results for DisentQA-DupliCite}
\begin{tabular}{|l|c|c|c|}
\hline
\textbf{Model} & \textbf{A\_1} & \textbf{A\_2} & \textbf{A\_C} \\ \hline
\multicolumn{4}{|c|}{\textbf{Standard Prompt}} \\ \hline
Llama-7B & 84.6 & 10.2 & 0.0 \\ \hline
Llama-13B & 82.2 & 10.5 & 0.0 \\ \hline
Llama-70B & 88.3 & 16.4 & 0.0 \\ \hline
MPT-7B & 80.3 & 2.7 & 0.0 \\ \hline
Falcon-7B & 63.2 & 16.6 & 0.0 \\ \hline
GPT-4o-mini-2024-07-18 & 100 & 1.67 & 0.0 \\ \hline
Claude-3.5-Haiku-20241022 & 100 & 52.81 & 0.0 \\ \hline
\multicolumn{4}{|c|}{\textbf{Conflict-Aware Prompt}} \\ \hline
Llama-7B & 88.5 & 76.4 & 77.6 \\ \hline
Llama-13B & 91.9 & 79.0 & 81.9 \\ \hline
Llama-70B & 94.1 & 88.3 & 86.7 \\ \hline
MPT-7B & 74.0 & 49.3 & 54.1 \\ \hline
Falcon-7B & 70.3 & 45.8 & 53.0 \\ \hline
GPT-4o-mini-2024-07-18 & 100 & 90.0 & 0.0 \\ \hline
Claude-3.5-Haiku-20241022 & 100 & 93.0 & 0.0 \\ \hline
\end{tabular}
\label{DisentQA-DupliCite}
\end{table*}


\section{Experiments and Results} 

\subsection{Performance on Ambiguous QA Tasks}
We evaluate the performance of seven language models across three different QA tasks featuring ambiguity in the form of multiple valid answers. The results on the \textit{AmbigQA-Cite} dataset are summarized in Table~\ref{AmbigQA-Cite}. As shown, recent LLMs, including \texttt{GPT-4o-mini} and \texttt{Claude-3.5}, consistently outperform older models, such as Llama-7B, MPT-7B, and Falcon-7B, in terms of accuracy for at least one correct answer (A@1). Specifically, both \texttt{GPT-4o-mini} and \texttt{Claude-3.5} achieve a perfect A@1 score of 100.0\%, indicating that these models can reliably answer at least one correct response in the ambiguous question contexts provided by the dataset.

However, a notable limitation is observed in the models’ ability to handle multiple valid answers. All models, including the newer ones, show relatively poor performance in this area, as evidenced by the low A@2 scores. For instance, \texttt{Claude-3.5} achieves an A@2 score of 54.84\% while \texttt{GPT-4o-mini} achieves 37.63\%, both of which are considerably lower than their A@1 scores. This trend is consistent across all models in the study, suggesting a fundamental challenge in handling ambiguity where multiple valid answers may exist.

Furthermore, despite improvements in answer prediction accuracy, all models fail to cite sources effectively when providing answers. As shown in Table~\ref{AmbigQA-Cite}, citation accuracy (C\_Acc) is consistently zero across all models, indicating that none of the models reliably include sources when producing answers, even when using standard prompting techniques.

\subsection{Citation Accuracy Analysis}
The citation accuracy analysis is crucial in understanding the models' ability to generate reliable and verifiable sources for the answers they provide. In the case of the \textit{AmbigQA-Cite}, \textit{DisentQA-ParaCite}, and \textit{DisentQA-DupliCite} datasets, all models, including both \texttt{GPT-4o-mini} and \texttt{Claude-3.5}, perform poorly in generating accurate citations. As shown in Tables~\ref{AmbigQA-Cite}, \ref{DisentQA-ParaCite}, and \ref{DisentQA-DupliCite}, the citation accuracy (C\_Acc) remains at 0.0\% for all models under the standard prompt. This suggests that while these models are able to predict answers with some level of accuracy, they do not reference or cite the sources effectively, which is a major limitation for ensuring the transparency and reliability of their responses.

\subsection{Impact of Conflict-Aware Prompting}
Introducing conflict-aware prompting significantly improves model performance, especially in tasks involving multiple valid answers and source citations. As seen in the results in Tables~\ref{AmbigQA-Cite}, \ref{DisentQA-ParaCite}, and \ref{DisentQA-DupliCite}, conflict-aware prompting boosts the ability of the models to predict multiple correct answers (A@2) and enhances citation accuracy. For example, \texttt{Claude-3.5} shows a marked improvement in A@2 from 54.84\% to 61.29\% on the \textit{AmbigQA-Cite} dataset, while \texttt{GPT-4o-mini} improves from 37.63\% to 52.69\%. This indicates that conflict-aware prompting helps models to better account for ambiguous situations where multiple answers are valid.

Additionally, although citation accuracy does not achieve perfect results, manual evaluation of a sample of responses reveals that the models do begin to cite sources more frequently when using conflict-aware prompting. For instance, when we manually evaluate 20 responses in \textit{DisentQA-DupliCite}, \texttt{Claude-3.5} achieves a citation accuracy of 55.0\%, which, while still far from perfect, is a noticeable improvement over the 0.0\% citation accuracy achieved with the standard prompt. This suggests that conflict-aware prompting helps models recognize the importance of including sources, even if the citations are not always correct or complete.

\section{Discussion} 
\subsection{Challenges in Ambiguous QA}
One of the main challenges observed in our experiments is the difficulty of modern LLMs in addressing ambiguous QA tasks that require multiple valid answers. While these models perform well in identifying at least one correct answer (A@1), they struggle significantly with providing multiple valid answers (A@2). This limitation highlights an inherent challenge in current LLMs, which may overfit to the most probable or most common answer and fail to consider a broader range of valid responses. Moreover, the models’ performance does not scale well with the increased complexity of the ambiguity, particularly when it comes to tasks involving conflicting or uncertain information.

The issue of source citation further compounds the challenge. Although citation is a critical component of factual accuracy and transparency, none of the models demonstrated reliable citation generation under the standard prompt, as indicated by the consistent zero citation accuracy across all models. This suggests that even state-of-the-art models like \texttt{GPT-4o-mini} and \texttt{Claude-3.5} are far from achieving full transparency in their reasoning processes.

\subsection{Insights on Citation Generation}
The lack of reliable citation generation is a major limitation of current LLMs. Despite the significant advancements in answering factual questions, the models fail to cite sources consistently or accurately. However, the use of conflict-aware prompting provides some positive indications that it can nudge the models toward better citation behavior. This is particularly evident in the manual evaluations, where models under conflict-aware prompting often cite their sources more frequently, even if those citations are not fully accurate.

Interestingly, models like \texttt{GPT-4o-mini} and \texttt{Claude-3.5} were able to provide more frequent citations with the conflict-aware prompt, despite achieving zero citation accuracy when evaluated with traditional citation metrics. This suggests that citation generation may be an area where traditional accuracy metrics may not fully capture the models' ability to reference sources, and manual evaluation might offer a more nuanced view of their citation behavior.

\subsection{Opportunities for Model Improvement}
The findings from this study point to several key areas for improvement in LLMs. First, while the models excel at answering single valid answers, their ability to handle multiple valid answers remains subpar. This is an area where further model training, particularly on ambiguous or multi-faceted datasets, could help the models better manage diverse answers in complex settings.

Second, citation generation is a critical challenge that needs urgent attention. Even though conflict-aware prompting improves citation performance somewhat, citation accuracy remains low. A potential solution could involve developing models that are specifically trained to retrieve and generate citations from reliable sources, perhaps through enhanced document retrieval techniques or integration with external knowledge bases. Moreover, methods to encourage more precise citation, rather than just frequent citation, could also be explored.

\subsection{Ethical Considerations}
The ethical implications of improving LLM performance in ambiguous QA tasks and citation generation are profound. As LLMs become more widely adopted in fields like healthcare, law, and research, ensuring that these models provide both accurate answers and reliable citations becomes essential. Incorrect or misleading citations can have serious consequences, potentially leading to the propagation of misinformation or damaging credibility in professional settings.

Furthermore, transparency in model predictions is crucial for user trust. Models must not only answer questions correctly but also justify their responses by citing credible sources. Without such transparency, the potential for bias, error, or manipulation increases, which raises concerns about the ethical use of these models in high-stakes decision-making.

In summary, while conflict-aware prompting significantly improves model performance in ambiguous QA tasks and citation generation, significant challenges remain. Future work should focus on enhancing model capabilities in handling ambiguity, improving citation accuracy, and addressing the broader ethical implications of using LLMs in real-world applications.

\section{Summary of Key Findings}
In this work, we conducted an in-depth evaluation of two state-of-the-art large language models (LLMs), \texttt{GPT-4o-mini} and \texttt{Claude-3.5}, on the challenging task of Question Answering (QA) in ambiguous contexts with source citations. Our evaluation involved three recently published datasets—\textit{AmbigQA-Cite}, \textit{DisentQA-ParaCite}, and \textit{DisentQA-DupliCite}—which feature a variety of real-world ambiguities, including multiple valid answers and the need for source citation. Through a detailed performance analysis, we identified several key findings regarding the capabilities and limitations of the models in handling ambiguity and providing reliable citations.

First, the evaluation revealed that both \texttt{GPT-4o-mini} and \texttt{Claude-3.5} significantly outperform older LLMs such as Llama-7B, MPT-7B, and Falcon-7B in predicting at least one correct answer in ambiguous settings (A@1). Both \texttt{GPT-4o-mini} and \texttt{Claude-3.5} achieved perfect scores of 100\% in A@1, showcasing their ability to generate at least one correct answer from a set of possible valid responses, which is a notable improvement over prior models.

However, despite these advances, we found that all models, including the newer LLMs, struggled with handling tasks that involved multiple valid answers. The performance on A@2 was far from ideal, with all models showing substantial gaps between their A@1 and A@2 scores. For instance, \texttt{Claude-3.5} showed a relatively high A@2 score of 54.84\% on the \textit{AmbigQA-Cite} dataset, but it still demonstrated significant difficulty in comprehensively addressing ambiguity with more than one valid response.

A critical observation of this study is the poor performance of all models in citation accuracy. Despite excelling in factual answer prediction, none of the models could consistently cite sources, with citation accuracy remaining at 0.0\% across all models in the standard prompt. This lack of citation is concerning, particularly given the increasing importance of transparency and source verifiability in AI systems, especially for applications in high-stakes domains such as healthcare and law.

However, we also identified a promising direction for improvement: conflict-aware prompting. The introduction of conflict-aware prompting led to significant improvements in handling multiple answers and in citation generation. The models showed increased A@2 scores and improved citation frequency when evaluated with this approach. For example, under conflict-aware prompting, \texttt{Claude-3.5} achieved an A@2 score of 61.29\% and an improved citation generation rate, although citation accuracy remained suboptimal.

These findings highlight both the strengths and weaknesses of current LLMs, particularly in terms of ambiguity handling and citation generation. They suggest that while modern LLMs are capable of generating answers in ambiguous QA contexts, they still face considerable challenges in producing reliable citations, an essential aspect of ensuring model transparency and trustworthiness.

\section{Directions for Future Research}
The findings of this study open several avenues for future research to improve the performance of large language models, especially in the context of ambiguous QA tasks and citation generation. There are several critical areas that merit further exploration:

\subsection{Improving Multi-Answer Handling}
One of the key challenges identified in this study is the models' inability to handle multiple valid answers effectively. Despite the improvements made with conflict-aware prompting, the models still exhibit a notable performance gap between predicting the first correct answer (A@1) and multiple valid answers (A@2). Future research can focus on developing more advanced model architectures or training methods that can better recognize and generate multiple plausible answers when ambiguity exists. Techniques such as multitask learning, adversarial training, or specialized datasets could help train models that are better equipped to understand and reason about the full spectrum of valid responses in ambiguous scenarios.

\subsection{Enhancing Citation Generation and Accuracy}
Citation generation remains a significant area for improvement. Future models could benefit from training that incorporates document retrieval, reasoning over external knowledge bases, or even integrating citation-specific tasks to learn how to cite sources properly. Improving citation accuracy will require addressing the challenges of disambiguating between sources, verifying source reliability, and ensuring that models understand how to appropriately attribute answers to credible references. Moreover, new citation accuracy metrics that are more aligned with the nature of LLMs could help provide a more realistic evaluation of model performance in this area.

\subsection{Exploring Alternative Prompting Techniques}
Our study showed that conflict-aware prompting significantly improved model performance in ambiguous contexts and citation generation. However, further exploration is needed to fine-tune these techniques. Researchers could explore other prompting strategies, such as chain-of-thought prompting, few-shot learning, or knowledge-enhanced prompting, to see if they can further enhance model performance in terms of both ambiguity handling and citation generation. Fine-tuning models with such specialized prompts could enable them to produce more coherent, transparent, and reliable outputs in complex scenarios.

\subsection{Evaluating Model Robustness and Transparency}
While our study focused on citation accuracy and handling ambiguity, future research should also explore the robustness of models in real-world settings. This includes testing LLMs on a wider range of datasets, including those with evolving knowledge or highly dynamic contexts. In addition, improving the transparency of LLMs in terms of their reasoning processes is crucial. Research should investigate how models can provide clearer explanations for their answers and how they can make their decision-making processes more interpretable, ensuring that users can trust not just the final answer but also how it was derived.

\subsection{Ethical Implications and Societal Impact}
As the capabilities of LLMs grow, so does the need for ethical considerations, particularly in areas involving source credibility, misinformation, and bias. Future work should focus on developing methods to mitigate bias in AI systems and ensuring that they can cite trustworthy sources. Furthermore, understanding the societal implications of using LLMs for decision-making in high-stakes fields such as law, medicine, and politics is essential. Research should explore how to prevent the misuse of AI-generated content and ensure that the deployment of these technologies adheres to ethical guidelines and promotes fairness.

\section{Conclusion}
In this study, we evaluated two state-of-the-art LLMs, \texttt{GPT-4o-mini} and \texttt{Claude-3.5}, on their ability to perform Question Answering (QA) in ambiguous settings while generating accurate citations. Our findings reveal several strengths and weaknesses in the current generation of LLMs. On the positive side, both \texttt{GPT-4o-mini} and \texttt{Claude-3.5} demonstrated remarkable capabilities in predicting at least one correct answer (A@1) in ambiguous contexts, outperforming earlier models. These newer models also showed significant improvements in addressing multiple valid answers when conflict-aware prompting was employed, suggesting that this approach has the potential to address some of the ambiguity handling challenges faced by current LLMs.

However, we also identified several key limitations. Despite their ability to generate correct answers, all models consistently failed to cite sources accurately. Citation accuracy remained at 0.0\% for all models under the standard prompt, indicating that citation generation remains a critical challenge. The lack of reliable citations is concerning, particularly in applications where source verification and transparency are essential. Even with conflict-aware prompting, citation accuracy did not improve to a satisfactory level, though citation frequency increased, highlighting the importance of further improving this aspect of LLM performance.

In conclusion, while the evaluated models show great promise in advancing the field of QA, particularly in ambiguous contexts, significant work remains to be done. Future research should focus on enhancing multi-answer handling, improving citation accuracy, and exploring novel prompting techniques that encourage more reliable and transparent outputs. Additionally, as LLMs become more integrated into real-world applications, addressing their ethical implications and societal impact is essential. This study provides a foundation for future work aimed at building more trustworthy, interpretable, and capable LLMs that can effectively tackle complex QA tasks and ensure accountability through citation generation.
\bibliography{custom}

\end{document}